\pdfoutput=1

\documentclass[11pt]{article}

\usepackage{acl}

\usepackage{times}
\usepackage{latexsym}

\usepackage[T1]{fontenc}

\usepackage[utf8]{inputenc}

\usepackage{microtype}

\usepackage{inconsolata}

\usepackage{booktabs} 
\usepackage{multirow}
\usepackage{enumitem}
\usepackage{caption}
\usepackage{subcaption}

\newcommand{\uhcomment}[1]{{\textcolor{cyan}{[Ulf: #1]}}}

\newcommand{\com}[1]{}

\title{Rediscovering Hashed Random Projections for Efficient Quantization of Contextualized Sentence Embeddings} 

\author{
  Ulf A. Hamster $^1$
  \quad
  Ji-Ung Lee $^2$
  \quad
  Alexander Geyken $^1$
  \quad
  Iryna Gurevych $^2$
  \\
  $^1$ Berlin-Brandenburgische Akademie der Wissenschaften, Berlin, Germany \\
  \texttt{\{hamster, geyken\}@bbaw.de} 
  \\
  $^2$ Ubiquitous Knowledge Processing Lab, Department of Computer Science \\ and Hessian Center for AI (hessian.AI), Technische Universität Darmstadt, Germany \\
  \texttt{\{lee, gurevych\}@ukp.informatik.tu-darmstadt.de}
}

\date{}

\usepackage{amsmath, amsfonts}
\usepackage{graphicx}
\usepackage{tabularx}

\usepackage{tikz}
\usetikzlibrary{positioning}

\begin{document}
\maketitle

\begin{abstract}
Training and inference on edge devices often requires an efficient setup due to computational limitations.
While pre-computing data representations and caching them on a server can mitigate extensive edge device computation, this leads to two challenges.
First, the amount of storage required on the server that scales linearly with the number of instances.
Second, the bandwidth required to send extensively large amounts of data to an edge device. 
To reduce the memory footprint of pre-computed data representations, we propose a simple, yet effective approach that uses randomly initialized hyperplane projections.
To further reduce their size by up to 98.96\%, we quantize the resulting floating-point representations into binary vectors.
Despite the greatly reduced size, we show that the embeddings remain effective for training models across various English and German sentence classification tasks that retain 94\%--99\% of their floating-point performance.



\end{abstract}

\section{Introduction}\label{sec:intro}

Neural networks have been subject to optimization attempts to reduce their memory requirement as well as computational cost~\citep{he-etal-2021-distiller, treviso2022efficient}.
Especially large language models (LLMs) such as BERT~\citep{devlin-etal-2019-bert} and GPT-3~\citep{brown2020language} require huge amounts of compute power and hence, cannot be run on edge devices (e.g., mobile phones, smart devices).
Whereas a possible solution is to simply increase the compute power on the server-side, this becomes quickly infeasible for training personalized models and results in a waste of resources as the compute power available on edge devices remains untapped.
To tackle this problem, various works focus on developing efficient language models (ELMs) that run on edge devices~\citep{sun-etal-2020-mobilebert, sankar-etal-2021-proformer, ge2022edgeformer, wang2022sparcl}.
However, their use cases primarily concern locally generated or small amounts of data (e.g., machine translation).  
Especially in cases like information retrieval, where most of the data remains static on the server, relying upon pre-computed representations such as sentence embeddings, is a viable approach~\citep{macavaney2020efficient,thakur2021beir}.
Such representations can then be sent to an edge device for training a local, personalized model.\footnote{Note, that this also increases privacy as a personalized model never leaves its device.}
However, a major bottleneck of using such pre-computed representations then becomes the transportation cost: as the size of the instances remains unaffected, sending a large amount of data to an edge device still requires a lot of bandwidth, while frequently sending small amounts results in a high number of requests made to the server.
Reducing the instance size hence becomes crucial in such cases as this allows us to send larger amounts at once for edge computation (cf. Table~\ref{tab:use-cases}).  

\begin{table}[]
    \centering
    \begin{small}
    \begin{tabular}{lccc}
        \toprule
         & Compute & Data & Device  \\
        \midrule
      LLMs (BERT, GPT, ...) & $\uparrow$ & $\uparrow$  & server \\ 
      LLMs (BERT, GPT, ...) & $\uparrow$ &  $\downarrow$ & server \\ 
      ELMs (MobileBERT, ...) & $\downarrow$ & $\downarrow$ & edge  \\ 
      This work & $\downarrow$ &  $\uparrow$ & edge \\ 
        \bottomrule
    \end{tabular}
    \end{small}
    \caption{Different scenarios of high ($\uparrow$) or low ($\downarrow$) available compute power and number of processed instances.}
    \label{tab:use-cases}
\end{table}

One method to effectively compress continuous representations into smaller, discrete ones is quantization~\citep{sheppard1897calculation}
It has been considered in NLP more recently; however, only with the goal of either compressing static, pre-trained word embeddings~\citep{liao2020embedding} or reducing the model size~\citep{zafrir2019q8bert, zadeh2020gobo, kuzmin2022fp, bai2022towards, dettmers2022gptint}.
In contrast to static embeddings such as GloVe~\citep{pennington-etal-2014-glove} that can be quantized in advance, contextual embeddings need to be quantized for each instance, leading to a computational overhead. 

To address the above problems, we propose to efficiently compress floating-point representations into binary ones by utilizing dimensionality reduction via hashed random projections (HRP) and quantization.
We show that we can reduce the memory footprint by up to 98.96\% with only a small degradation in performance.
Our experiments using six multi-lingual models on seven English and four German sentence classification tasks, as well as five English semantic similarity datasets show that HRP perform well across different projection sizes, which allows practitioners to adapt them to suit their available memory.

\section{Use Case}\label{sec:use-case}
\begin{figure}[!ht]
        \centering
        \includegraphics[width=.41\textwidth]{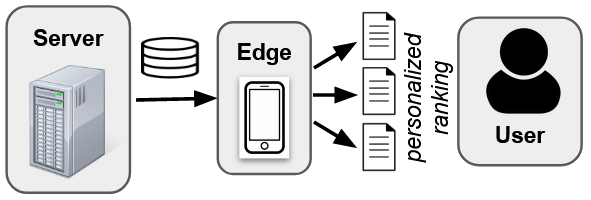}
        \caption{Edge device training of a personalized ranker.} 
        \label{fig:use-case}
\end{figure}

Figure~\ref{fig:use-case} shows the general use case we target; namely, interactively training a personalized ranking model for larger batches of data.
The whole data (e.g., large corpora of several hundreds GB) is stored on the server in a database.
To interactively train a model with each choice of the user, it is key to have an efficient model running on the edge device to ensure a low overall latency.
As frequently sending small batches of data increases power consumption on the device with each request made to the server, we also want to pre-load as much data as possible.
To maximize the number of pre-loaded instances, it is thus crucial that each instance is as small as possible.
Moreover, as the data stored on the server remains static, we can pre-compute its representation to additionally reduce the computational burden on the edge device.

\section{Approach}\label{sec:approach}
\citet{indyk1998} as well as \citet{charikar2002} have shown that hashes from randomly initialized hyperplane projections can approximate the hamming distance in nearest neighbor clustering as well as similarity functions in local-sensitivity hashing.
Following their idea, we randomly initialize a hyperplane $W_{\mathrm{hrp}} \in \mathbb{R}^{d_s \times d_t} \sim \mathcal{N}(0,1)$ to project a sentence embedding of dimension $d_s$ into a target dimension $d_t$.
We then further apply the Heaviside~\cite{weisstein2002heaviside} step function $H(\cdot)$ to quantize the resulting vector.
This allows us to transform any floating-point vector $x_{\mathrm{SBert}} \in \mathbb{R}^{d_s}$ into a binary vector $x^{0/1}_{\mathrm{SBert}} \in \mathbb{B}^{d_t}$.
Equation (1) and (2) show the definition of the Heaviside step function and the whole operation, respectively.
\begin{align}
H(z) &= 
\begin{cases}
  \textsc{1}  & \text{if } z \geq 0 \\
  \textsc{0} & \text{else}
\end{cases}
\\
x^{0/1}_{\mathrm{SBert}} &= H\left(x_{\mathrm{SBert}} \cdot W_{\mathrm{hrp}} \right)
\end{align}

\section{Experiments}\label{sec:experiments}
We focus on six multilingual sentence embedding models and evaluate them across twelve English and four German downstream tasks.

\subsection{Datasets}
Considering our scenario to train a personalized ranking model for sentences, we focus on sentence-level tasks.
We evaluate our approach across five English binary (namely, CR, MR, MPQA, MRPC, and SUBJ) and two multi-class (SST5 and TREC) sentence classification tasks using SentEval~\citep{conneau2018}. 
We further evaluate how well the projected embeddings preserve the semantic textual similarity (STS) on the STS\{12--16\} tasks on English texts that are also available in SentEval.
For German, we evaluate two binary (VMWE, MIO-P) and two multi-class (ABSD-2, ARCH) sentence classification tasks.\footnote{We provide more detailed descriptions of the datasets in the appendix~\ref{sec:appendix-datasets}.} 

\subsection{Experimental Setup}
\label{sec:experimental-setup}
We consider various embedding models and HRP output dimensions for our experiments. 
Our baseline is the sigmoid quantization method proposed by \citet{grzegorczyk2017}.
As we aim to evaluate the performance of the HRP layers with quantization compared to the original representations, we train a simple linear classification model that uses the quantized vectors as inputs and outputs a vector respective to the number of classes passed into the softmax function for a prediction. 
For the sentence similarity tasks, we directly compare the resulting vectors after quantization.

\paragraph{Sentence representations.}
Our evaluated models involve six pre-trained sentence embedding models that support both English and German text:

\begin{description}[topsep=5pt,itemsep=2pt]
    \item \textbf{SBert MPnet v2}: the \textit{paraphrase-multilingual-mpnet-base-v2} model provided by sentence BERT~(SBert; \citealt{reimers2019}) with an output dimension of 768.
    \item \textbf{MiniLM v2}: the \textit{paraphrase-MiniLM-L12-v2} SBert model with an output dimension of 384.
    \item \textbf{DistilUSE v2}: the \textit{distiluse-base-multilingual-cased-v2} SBert model with an output dimension of 512.
    \item \textbf{LaBSE}: a model trained by \citet{feng2022} using the SBert framework with an output dimension of 768.
    \item \textbf{m-USE}: a multilingual semantic retrieval model~\citep{yang2020} with an output dimension of 512.
    \item \textbf{LASER}: language agnostic sentence representations developed by \citet{artetxe2019} with an output dimension of 1024. This is the only model that requires us to specify the language of the inputs. 
\end{description}


\paragraph{HRP dimensions.}
The outputs of each embedding model are compressed with HRP layers of different dimensions (256 384, 512, 768, 1024, 1536, 2048) which is also the bit-size of the hashed projection.
Given an embedding model outputs a 32-bit floating-point of $d_o$ vector elements, and the HRP layer of $d_h$ bits, the memory consumption rate is $\frac{d_h}{32\cdot d_o}$.
In case of the sigmoid baseline compression model, the memory consumption rate is always $\frac{d_o}{32\cdot d_o}=\frac{1}{32}$.

\paragraph{Hyperparameters.}
For SentEval classification tasks, we use RMSProp~\citep{tieleman2012lecture}, batch size 128,  tenacity 3, and train for 2 epochs in a 5-fold cross validation.
For the German data, we use the pre-defined data splits with AMSGrad~\citep{reddi2018}, 
batch size 128, and train for 500 epochs.\footnote{See PyPi package \href{https://pypi.org/project/sentence-embedding-evaluation-german/0.1.9/}{sentence-embedding-evaluation-german, version 0.1.9}}
For each embedding model and HRP dimension, we run 10 different weight initializations using different pseudo-random number generator seeds.
All experiments were conducted on a server with 64 CPU cores with 504GB RAM and one NVIDIA A100 and took $\sim$18 hours.

\begin{figure}[!ht]
    \begin{subfigure}[b]{0.49\textwidth}
        \centering
        \includegraphics[width=.95\textwidth]{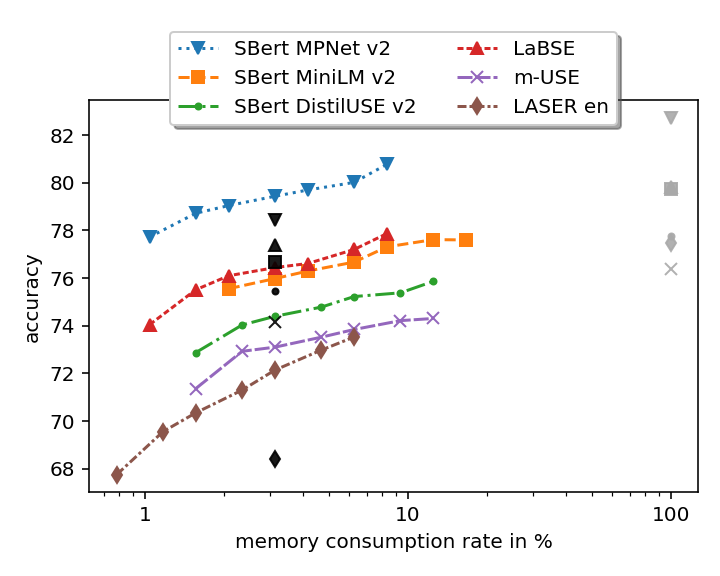}
        \vspace{-1em}
        \caption{Average accuracy for English sentence classification tasks.} 
        \label{fig:avg-senteval}
    \end{subfigure}
    \hfill 
    \begin{subfigure}[b]{0.49\textwidth}
        \centering
        \includegraphics[width=.95\textwidth]{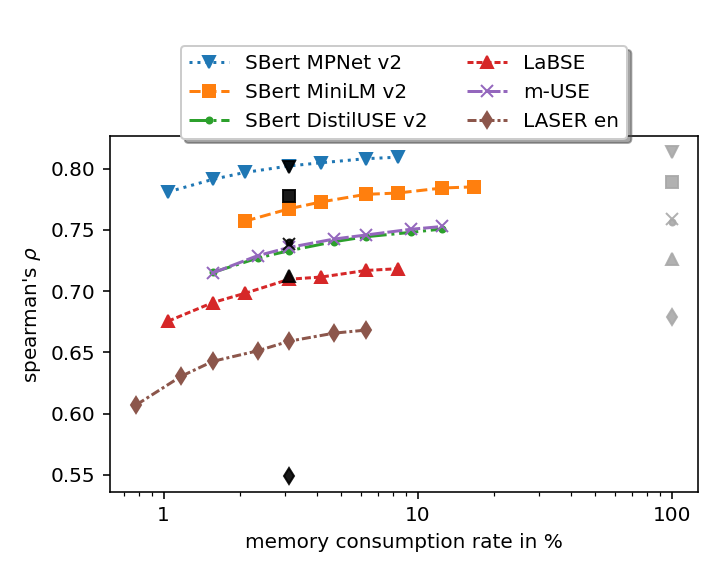}    
        \vspace{-1em}
        \caption{Average Spearman's $\rho$ for English sentence similarity tasks.} 
        \label{fig:avg-similarity}
    \end{subfigure}
    \hfill
    \begin{subfigure}[b]{0.49\textwidth}
        \centering
        \includegraphics[width=.95\textwidth]{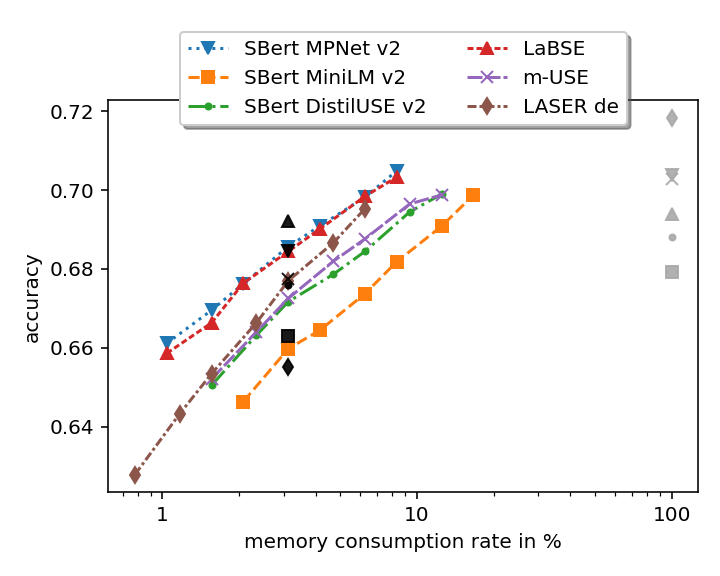}
        \vspace{-1em}
        \caption{Average accuracy for German sentence classification tasks.} 
        \label{fig:avg-seeg}
    \end{subfigure}
    \caption{
        Experimental results. 32-bit floating-point results are in grey and the sigmoid baseline in black. }
\end{figure}

\paragraph{Results.}
Figures~\ref{fig:avg-senteval}--\ref{fig:avg-seeg} show the performance of all models averaged across all tasks and runs.\footnote{We provide more detailed results in appendix~\ref{sec:appendix-results}.}
Interestingly, all models express a similar trend with respect to their performance changes compared to the target dimension (i.e., the memory consumption rate). 
Moreover, we find that decreasing the target dimension of the HRP projections has a substantially smaller impact on the sentence similarity tasks (Figure~\ref{fig:avg-similarity}).
For instance, we can see that the target dimension of 256 for the best performing model (SBert MPNet v2) only results in an average decrease of 3.26 Spearman's $\rho$ compared to using the 32-bit floating-point representation (i.e., using only 1.04\% of the initial size).
Furthermore, We can  observe that the German representations are more sensitive to reducing the target dimension and show a steeper decline.
Finally, we find that SBert models still perform surprisingly well, especially for sentence similarity tasks, outperforming more recent models such as SimCSE~\citep{gao-etal-2021-simcse} and CPT$_\mathrm{XL}$~\citep{neelakantan2022text}.

%

\section{Related Work}\label{sec:rel-work}
Most works focus on developing \textit{edge computation models}, i.e., models that can be utilized on edge devices.
In contrast, our goal is to directly compress the representations (\textit{embedding compression}) to increase storage efficiency and reduce bandwidth. 

\paragraph{Edge computation models.}
While approaches such as knowledge distillation~\citep{hinton2015} and pruning~\citep{lecun1990} can generally reduce a model's size~\citep{sanh2020distil, rolfs2021}, developing well-performing models for edge computing often requires additional measures.
For instance, \citet{sun-etal-2020-mobilebert} show that using a bottleneck architecture together with various architectural changes can achieve a comparable performance to BERT-base with only a quarter of the model parameters.
Other works investigate various kinds of sparsity (e.g., in the attention mechanism) to decrease the number of model weights to substantially reduce the model size~\citep{ge2022edgeformer, wang2022sparcl}.
Finally, one line of work investigates the use of quantization or mixed-precision weights and training to decrease model size and improve run-time efficiency~\citep{micikevicius2018mixed, bai2022towards, dettmers2022gptint, kuzmin2022fp, zafrir2019q8bert}.
However, such approaches are often dependent on the underlying hardware and hence difficult to utilize in edge devices on a broad scale, as they comprise various types of hardware. 
Close to our approach are the works by \citet{ravi2019efficient} and \citet{sankar-etal-2021-proformer} who investigate utilizing projection layers at various stages---however, they still rely on an end-to-end model training.
Hence, the above approaches do not affect the resulting representations, which make them infeasible for scenarios where large amounts of instances are sent to an edge device.

\paragraph{Embedding compression.}
Several works investigate reducing the memory footprint of dense word representations such as GloVe~\citep{pennington-etal-2014-glove} and word2vec~\citep{mikolov2013distributed}.
For instance, \citet{shu2018compressing} and \citet{kim-etal-2020-adaptive} propose to utilize code-book approaches that transform continuous vectors into discrete ones.
Others propose to maximize the bit variance~\citep{mu2018allbutthetop} or isotropy~\citep{liao2020embedding}.
Interestingly, both approaches show that they can substantially reduce the memory footprint while preserving high performance even though they pursue opposite objectives~\citep{liao2020embedding}.
This may indicate that the models are capable of recovering the information loss resulting from quantization as long as the projection method remains the same.
A drawback of the above approaches is that they require training which would lead to a substantial computational overhead for contextualized embeddings, making them infeasible, especially in large-scale retrieval setups.

Closest to this work are three works that also investigate hashed random projections with images~\citep{choromanska2016binary}, bag-of-word representations~\citep{wojcik2018random}, and static word-embeddings~\citep{grzegorczyk2017}.
One substantial difficulty that \citet{wojcik2018random} and \citet{grzegorczyk2017} identify is the aggregation of quantized, static word embeddings into paragraph or document embeddings; which motivates them to tackle this issue by additionally training the projection layer.
While this is easily possible for static embeddings with a fixed-size vocabulary, it quickly becomes infeasible with contextualized word and sentence embeddings.
In contrast, we find that recent aggregation methods for contextualized word embeddings such as SBert~\citep{reimers2019} can work well with HRP and a minimal computational overhead.

\section{Conclusion}\label{sec:conclusion}

In this work, we explored HRP and quantization to compress floating-point sentence embeddings into binary vectors.
Our results indicate that much of the existing information in sentence embeddings can be discarded as even simple models retain up to 99\% of their original performance.
We plan to investigate this more closely in future work to identify what exactly makes these embedding models work so well and whether floating-point representations are even necessary at all to capture semantic similarity on a sentence level.
We conclude that HRP is a promising approach for retrieval-like tasks that rely on semantic similarity.

\section*{Impact Statement}\label{sec:impact}
\paragraph{Limitations.}
Although our experiments show that HRP can substantially reduce the memory footprint with only a minor impact on the performance, we find several limitations that remain to be investigated in future works.
As expected, we find that a higher target dimension leads to a better result.
Consequently, the problem ultimately becomes a choice to be made by a developer considering their explicit use case and their limits on the bandwidth and memory consumption.
Second, our fixed projection matrix is initialized randomly, which may cause crucial dimensions to be omitted. 
A more informed approach using dimensional reduction methods could lead to better projections.


\section*{Acknowledgments}
We thank Gregor Middel for pointing out the deployment problems, and Jan-Christoph Klie, Indraneil Paul, Dennis Zyska and Chen Liu for their helpful feedback.
The Evidence project was funded by the Deutsche Forschungsgemeinschaft (DFG, German Research Foundation) - 433249742 (GU 798/27-1; GE 1119/11-1).

\bibliography{acl2021}

\clearpage
\newpage

\appendix
\section{Appendices}
\label{sec:appendix}

\subsection{Detailed Description of Datasets}\label{sec:appendix-datasets}
For evaluation, we use the following binary (marked by $0/1$) and multi-class classification tasks (with five and six classes, respectively).

\begin{description}[topsep=5pt,itemsep=3pt]
    \item[CR$^{0/1}$] Sentiments classification of customer product reviews \citep{hu2004}.
    \item[MR$^{0/1}$] Sentiment classification of movie reviews \citep{pang2005}.
    \item[MPQA$^{0/1}$] Polarity classification of opinions \citep{wiebe2005}.
    \item[MRPC$^{0/1}$] Paraphrase detection \citep{dolan2004}. 
    \item[SUBJ$^{0/1}$] Subjectivity versus objectivity classification \citep{pang2004}.
    \item[SST5$^{5}$] Sentiment classification of movie reviews \citep{socher2013}.
    \item[TREC$^{6}$] Question type classification \citep{li2002}.
\end{description}
The German sentence classification tasks:
\begin{description}[topsep=5pt,itemsep=3pt]
    \item[VMWE$^{0/1}$] Classify sentences with multiword expressions (MWEs) if these are figuratively or literally. Texts are in Standard German (lang:de) from a newspaper corpus \citep{ehren-etal-2020-supervised}.
    \item[ABSD-2$^{3}$] Sentiment classification of customer feedback (synchronic). Texts are in Standard German (lang:de) \citep{germevaltask2017}.
    \item[MIO-P$^{0/1}$] Binary classification if blog comment contains a personal story. Texts are Standard German in Austria (lang: de-AT) or the Austro-Bavarian dialect (lang:bar) \citep{Schabus2017}.
    \item[ARCH$^{4}$] Swiss dialect classification (lang:gsw) based on transcribed audio recordings \citep{samardzic-etal-2016-archimob}.
\end{description}

\subsection{Model architecture for the SentEval tasks}
\begin{figure}[ht!]
\centering

\begin{tikzpicture}[x=0.75pt,y=0.75pt,yscale=-1,xscale=1]

\draw    (59.75,96.75) -- (59.75,79.5) ;
\draw [shift={(59.75,76.5)}, rotate = 90] [fill={rgb, 255:red, 0; green, 0; blue, 0 }  ][line width=0.08]  [draw opacity=0] (8.04,-3.86) -- (0,0) -- (8.04,3.86) -- (5.34,0) -- cycle    ;
\draw   (29.5,96.5) -- (89,96.5) -- (89,116) -- (29.5,116) -- cycle ;
\draw   (29.5,56.17) -- (89.25,56.17) -- (89.25,75.67) -- (29.5,75.67) -- cycle ;
\draw   (29.58,15.42) -- (89.25,15.42) -- (89.25,34.92) -- (29.58,34.92) -- cycle ;
\draw    (59.25,56.25) -- (59.25,39) ;
\draw [shift={(59.25,36)}, rotate = 90] [fill={rgb, 255:red, 0; green, 0; blue, 0 }  ][line width=0.08]  [draw opacity=0] (8.04,-3.86) -- (0,0) -- (8.04,3.86) -- (5.34,0) -- cycle    ;
\draw    (59.5,136.75) -- (59.5,119.5) ;
\draw [shift={(59.5,116.5)}, rotate = 90] [fill={rgb, 255:red, 0; green, 0; blue, 0 }  ][line width=0.08]  [draw opacity=0] (8.04,-3.86) -- (0,0) -- (8.04,3.86) -- (5.34,0) -- cycle    ;
\draw    (59.15,15.15) -- (59.15,3.8) ;
\draw [shift={(59.15,0.8)}, rotate = 90] [fill={rgb, 255:red, 0; green, 0; blue, 0 }  ][line width=0.08]  [draw opacity=0] (8.04,-3.86) -- (0,0) -- (8.04,3.86) -- (5.34,0) -- cycle    ;
\draw  [dash pattern={on 4.5pt off 4.5pt}]  (140.67,48) -- (203.67,48) ;
\draw  [dash pattern={on 4.5pt off 4.5pt}]  (4,48) -- (54.67,48) ;
\draw    (197.67,48) -- (197.67,35.67) ;
\draw [shift={(197.67,33.67)}, rotate = 90] [color={rgb, 255:red, 0; green, 0; blue, 0 }  ][line width=0.75]    (4.37,-1.32) .. controls (2.78,-0.56) and (1.32,-0.12) .. (0,0) .. controls (1.32,0.12) and (2.78,0.56) .. (4.37,1.32)   ;
\draw    (197.67,62.33) -- (197.67,50) ;
\draw [shift={(197.67,48)}, rotate = 90] [color={rgb, 255:red, 0; green, 0; blue, 0 }  ][line width=0.75]    (4.37,-1.96) .. controls (2.78,-0.92) and (1.32,-0.27) .. (0,0) .. controls (1.32,0.27) and (2.78,0.92) .. (4.37,1.96)   ;

\draw (59.63,106.34) node  [font=\scriptsize] [align=left] {SBert};
\draw (60.69,66.17) node  [font=\tiny] [align=left] {Hashed Random \\Projection (HRP)};
\draw (60.42,25.17) node  [font=\tiny] [align=left] {\begin{minipage}[lt]{31.35pt}\setlength\topsep{0pt}
\begin{center}
Linear Model
\end{center}

\end{minipage}};
\draw (65.75,122.75) node [anchor=north west][inner sep=0.75pt]  [font=\scriptsize] [align=left] {input sentence};
\draw (68,78.65) node [anchor=north west][inner sep=0.75pt]  [font=\scriptsize]  {$x_{\mathrm{SBert}} \in \mathbb{R}^{d_{s}}$};
\draw (70,39.4) node [anchor=north west][inner sep=0.75pt]  [font=\scriptsize]  {$x^{0/1}_{\mathrm{SBert}} \in \mathbb{R}^{d_{t}}$};
\draw (110,65.37) node [anchor=west] [inner sep=0.75pt]  [font=\scriptsize]  {$W_{\mathrm{hrp}} \in \mathbb{R}^{d_{s} \times d_{t}}$};
\draw (197.26,65.3) node [anchor=east] [inner sep=0.75pt]  [font=\tiny,rotate=-270] [align=left] {preprocessing};
\draw (196.93,29.92) node [anchor=west] [inner sep=0.75pt]  [font=\tiny,rotate=-270] [align=left] {training};

\end{tikzpicture}
\caption{
    Feature extraction with pre-trained SBert, embedding compression with HRP layer before the SentEval/SEEG task.
}
\label{fig:senteval-hrp-model}
\end{figure}
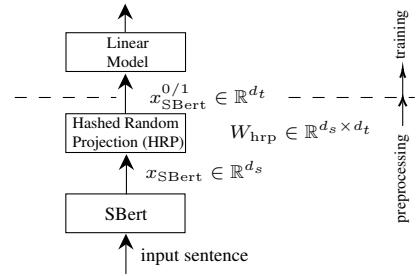

Figure~\ref{fig:senteval-hrp-model} shows the whole pipeline, consisting of HRP embedding compression (during preprocessing) and SentEval/SEEG training. 

\subsection{Software Contributions}\label{sec:software}
\begin{itemize}
\item
The computational experiments are available on \href{https://github.com/satzbeleg/study-hrp}{https://github.com/satzbeleg/study-hrp}
\item
The Hashed Random Projection (HRP) layer for Keras/TF2 is available on 
\href{https://github.com/satzbeleg/keras-hrp}{https://github.com/satzbeleg/keras-hrp} (GitHub)
and 
\href{https://pypi.org/project/keras-hrp/ }{https://pypi.org/project/keras-hrp/} (PyPi)
\item
The code for German sentence embedding evaluation is available 
on
\href{https://github.com/ulf1/sentence-embedding-evaluation-german}{https://github.com/ulf1/sentence-embedding-evaluation-german}
(GitHub) 
and 
\href{https://pypi.org/project/sentence-embedding-evaluation-german/}{https://pypi.org/project/sentence-embedding-evaluation-german/} (PyPi)
\end{itemize}

\subsection{Detailed results}\label{sec:appendix-results}
Table~\ref{tab:seeg-downstream}--\ref{tab:senteval-sts} show all results across all models and projection dimensions for all sentence classification and semantic similarity tasks, respectively.

\begin{table*}[!htb]
\centering \small
\begin{tabular}{rllllllll}
\toprule
Model/ target dimension & MR & CR & SUBJ & MPQA & SST5 & TREC & MRPC & Avg. \\
\midrule
CPT$_{\mathrm{XL}}$~\citep{neelakantan2022text} & \textbf{92.4} & \textbf{93.9} & \textbf{97.0} & \textbf{91.8} & - & \textbf{96.4} & \textbf{78.1} & \textbf{84.9} \\
\midrule
SBert MPNet v2 & 83.57 & 91.71 & 94.27 & 90.64 & \textbf{51.13} & 91.20 & 76.46 & 82.71 \\
sigmoid & 81.93 & 90.14 & 92.46 & 89.47 & 44.39 & 83.00 & 67.77 & 78.45 \\
256 & 80.21 & 89.39 & 89.91 & 88.09 & 46.09 & 79.14 & 71.25 & 77.72 \\
384 & 80.94 & 89.76 & 91.00 & 88.78 & 46.15 & 82.04 & 72.39 & 78.72 \\
512 & 81.63 & 90.05 & 91.72 & 89.03 & 47.35 & 81.72 & 71.79 & 79.04 \\
768 & 81.95 & 90.44 & 91.97 & 89.38 & 45.92 & 83.40 & 72.98 & 79.43 \\
1024 & 81.65 & 90.52 & 92.53 & 89.71 & 47.24 & 84.44 & 71.80 & 79.70 \\
1536 & 82.29 & 90.73 & 92.74 & 89.95 & 46.91 & 85.64 & 71.85 & 80.02 \\
2048 & 82.25 & 90.74 & 92.86 & 90.02 & 48.04 & 87.24 & 74.29 & 80.78 \\
\midrule 
SBert MiniLM v2 & 79.59 & 88.42 & 92.20 & 88.95 & 45.20 & 89.40 & 74.49 & 79.75 \\
sigmoid & 76.22 & 87.18 & 90.73 & 88.27 & 42.62 & 81.80 & 70.03 & 76.69 \\
256 & 75.37 & 86.27 & 88.26 & 87.03 & 42.63 & 77.84 & 71.48 & 75.56 \\
384 & 76.05 & 86.41 & 89.56 & 87.72 & 42.34 & 78.02 & 71.76 & 75.98 \\
512 & 76.53 & 86.83 & 89.93 & 87.89 & 43.00 & 80.20 & 69.71 & 76.30 \\
768 & 77.21 & 87.48 & 90.57 & 88.00 & 42.41 & 81.40 & 69.65 & 76.67 \\
1024 & 77.11 & 87.16 & 90.78 & 88.22 & 43.26 & 83.12 & 71.53 & 77.31 \\
1536 & 77.71 & 87.66 & 91.14 & 88.60 & 43.36 & 82.66 & 72.17 & 77.62 \\
2048 & 77.46 & 87.64 & 91.23 & 88.69 & 43.33 & 82.18 & 72.68 & 77.60 \\
\midrule 
SBert DistilUSE v2 & 75.01 & 83.34 & 91.89 & 87.52 & 43.89 & 92.60 & 70.14 & 77.77 \\
sigmoid & 71.64 & 81.14 & 89.83 & 85.93 & 40.14 & 87.80 & 71.71 & 75.46 \\
256 & 68.16 & 77.74 & 87.34 & 84.60 & 38.09 & 83.98 & 70.23 & 72.88 \\
384 & 69.91 & 78.84 & 88.75 & 85.39 & 38.79 & 86.62 & 69.97 & 74.04 \\
512 & 69.85 & 79.17 & 89.41 & 85.68 & 38.60 & 86.96 & 71.11 & 74.40 \\
768 & 71.26 & 79.72 & 89.83 & 86.16 & 39.37 & 86.74 & 70.40 & 74.78 \\
1024 & 71.10 & 80.21 & 90.35 & 86.05 & 39.34 & 88.42 & 71.12 & 75.23 \\
1536 & 71.91 & 80.71 & 90.74 & 86.45 & 38.09 & 88.32 & 71.46 & 75.38 \\
2048 & 72.11 & 80.91 & 90.88 & 86.71 & 41.38 & 89.18 & 69.88 & 75.86 \\
\midrule 
LaBSE & 78.81 & 86.25 & 92.76 & 89.66 & 46.61 & 90.60 & 73.97 & 79.81 \\
sigmoid & 74.85 & 84.32 & 91.38 & 88.32 & 39.32 & 90.20 & 73.39 & 77.40 \\
256 & 71.46 & 80.36 & 88.41 & 86.03 & 39.43 & 81.76 & 70.88 & 74.05 \\
384 & 72.37 & 81.81 & 89.58 & 86.85 & 41.75 & 84.46 & 71.78 & 75.51 \\
512 & 73.84 & 82.33 & 90.08 & 87.59 & 41.05 & 86.00 & 71.81 & 76.10 \\
768 & 74.50 & 83.47 & 90.92 & 87.82 & 42.04 & 85.58 & 70.78 & 76.44 \\
1024 & 75.04 & 84.11 & 91.19 & 87.91 & 41.78 & 85.82 & 70.44 & 76.61 \\
1536 & 75.76 & 84.58 & 91.43 & 88.58 & 42.45 & 87.84 & 69.77 & 77.20 \\
2048 & 75.97 & 84.52 & 91.83 & 88.59 & 43.37 & 88.50 & 72.31 & 77.87 \\
\midrule 
m-USE & 72.39 & 79.97 & 91.03 & 87.57 & 41.81 & 92.20 & 69.80 & 76.40 \\
sigmoid & 69.05 & 77.32 & 89.24 & 85.73 & 37.65 & 87.20 & 73.04 & 74.18 \\
256 & 66.02 & 75.15 & 87.01 & 84.58 & 35.74 & 82.16 & 68.92 & 71.37 \\
384 & 66.90 & 76.03 & 88.11 & 85.14 & 36.84 & 85.86 & 71.63 & 72.93 \\
512 & 67.68 & 77.03 & 88.46 & 85.68 & 36.61 & 85.82 & 70.44 & 73.10 \\
768 & 68.33 & 76.89 & 89.07 & 85.74 & 36.66 & 87.14 & 70.82 & 73.52 \\
1024 & 68.48 & 77.84 & 89.52 & 86.24 & 37.51 & 87.66 & 69.66 & 73.84 \\
1536 & 68.99 & 77.82 & 89.81 & 86.26 & 37.37 & 88.48 & 70.80 & 74.22 \\
2048 & 68.59 & 78.14 & 90.10 & 86.46 & 37.96 & 89.50 & 69.39 & 74.31 \\
\midrule 
LASER en & 74.08 & 79.73 & 91.48 & 88.38 & 44.25 & 89.20 & 75.19 & 77.47 \\
sigmoid & 58.98 & 68.98 & 79.97 & 85.77 & 33.48 & 80.40 & 71.19 & 68.40 \\
256 & 63.04 & 70.97 & 81.20 & 82.42 & 34.17 & 73.44 & 69.09 & 67.76 \\
384 & 63.65 & 71.71 & 82.89 & 84.36 & 34.98 & 78.78 & 70.57 & 69.56 \\
512 & 64.79 & 73.02 & 84.71 & 84.65 & 36.03 & 79.00 & 70.25 & 70.35 \\
768 & 66.08 & 73.36 & 85.78 & 85.45 & 36.77 & 80.80 & 70.90 & 71.30 \\
1024 & 66.39 & 75.09 & 86.99 & 86.41 & 37.07 & 82.64 & 70.42 & 72.14 \\
1536 & 68.00 & 75.55 & 88.12 & 86.53 & 37.93 & 83.54 & 71.21 & 72.98 \\
2048 & 68.02 & 76.67 & 88.83 & 87.03 & 37.80 & 84.74 & 71.61 & 73.53 \\
\bottomrule
\end{tabular}
\caption{Avg. Accuracy score for SBert with HRP-layers on English SentEval classification tasks. The reported scores are the average of the 10 runs with random HRP initializations. CPT$_{\mathrm{XL}}$ reports the scores of a recent state-of-the-art model.} 
\label{tab:senteval-downstream}
\end{table*}

\begin{figure*}[!ht]
    \begin{subfigure}[b]{0.49\textwidth}
        \centering
        \includegraphics[width=.9\textwidth]{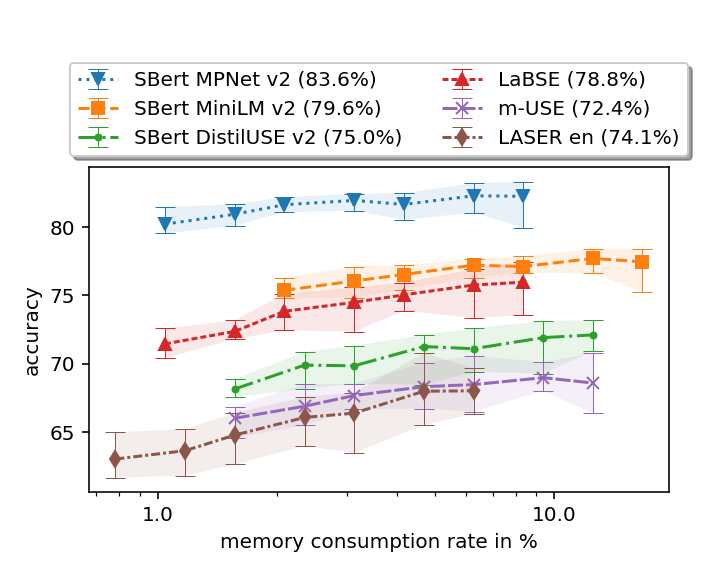}
        \vspace{-1em}
        \caption{MR.} 
        \label{fig:detail-mr}
    \end{subfigure}
    \hfill
    \begin{subfigure}[b]{0.49\textwidth}
        \centering
        \includegraphics[width=.9\textwidth]{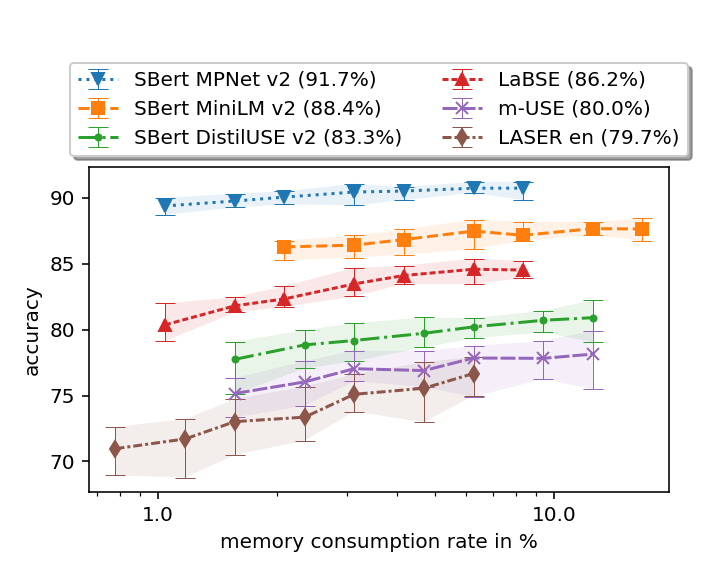}
        \vspace{-1em}
        \caption{CR.} 
        \label{fig:detail-cr}
    \end{subfigure}
    \hfill
    \begin{subfigure}[b]{0.49\textwidth}
        \centering
        \includegraphics[width=.9\textwidth]{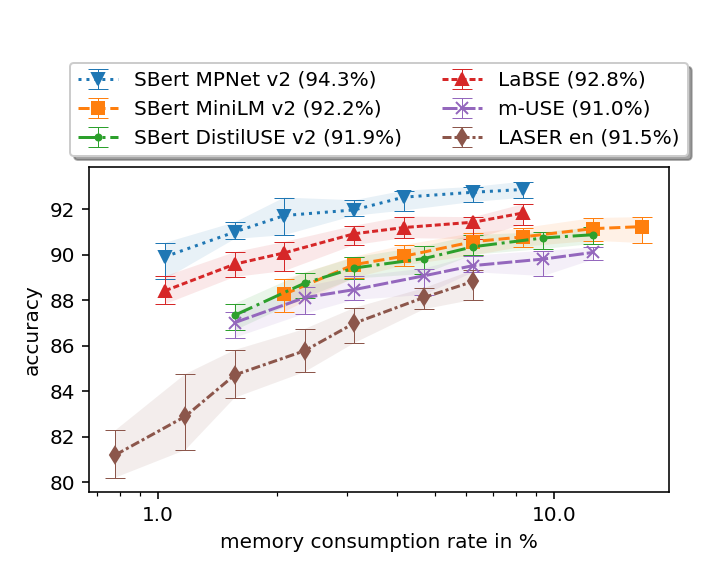}
        \vspace{-1em}
        \caption{SUBJ.} 
        \label{fig:detail-subj}
    \end{subfigure}
    \hfill
    \begin{subfigure}[b]{0.49\textwidth}
        \centering
        \includegraphics[width=.9\textwidth]{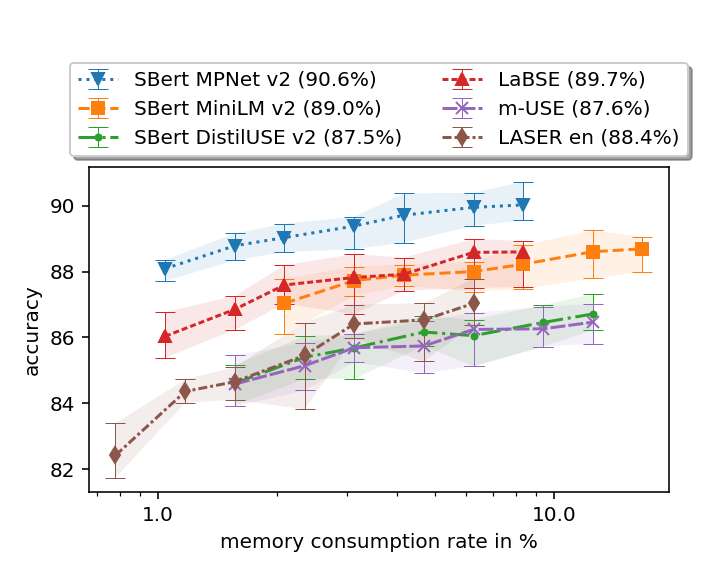}
        \vspace{-1em}
        \caption{MPQA.} 
        \label{fig:detail-mpqa}
    \end{subfigure}
    \hfill
    \begin{subfigure}[b]{0.49\textwidth}
        \centering
        \includegraphics[width=.9\textwidth]{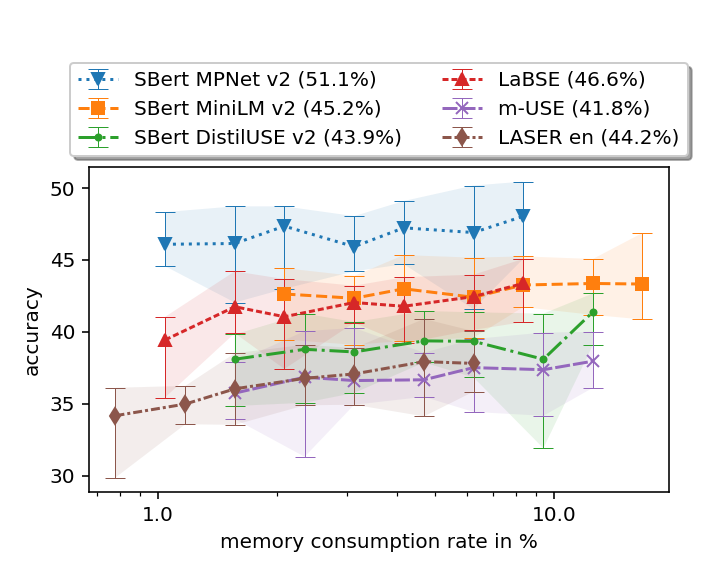}
        \vspace{-1em}
        \caption{SST5.} 
        \label{fig:detail-sst5}
    \end{subfigure}
    \hfill
    \begin{subfigure}[b]{0.49\textwidth}
        \centering
        \includegraphics[width=.9\textwidth]{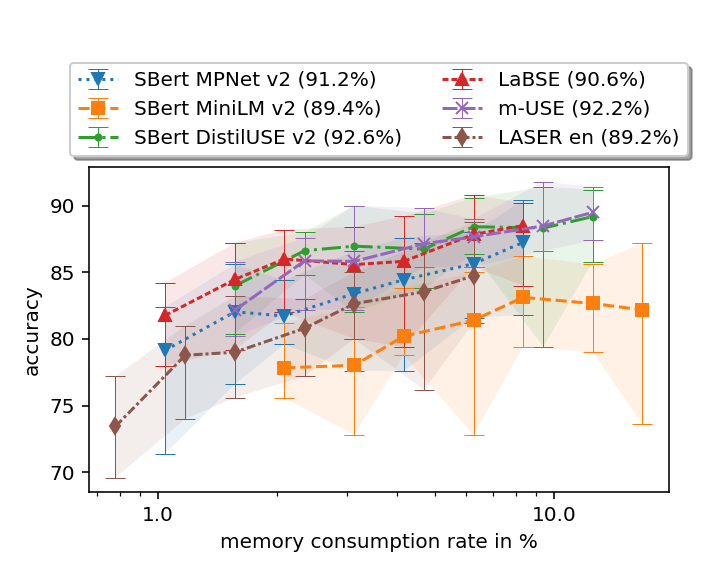}
        \vspace{-1em}
        \caption{TREC.} 
        \label{fig:detail-trec}
    \end{subfigure}
    \hfill
    \begin{subfigure}[b]{0.49\textwidth}
        \centering
        \includegraphics[width=.9\textwidth]{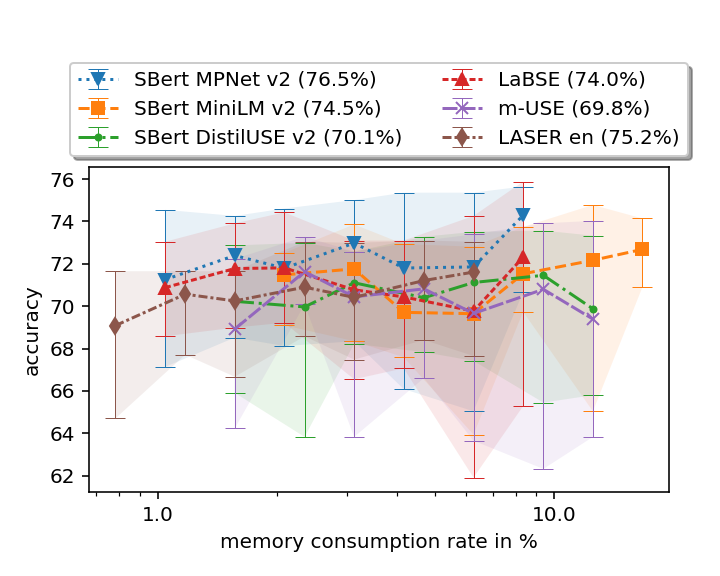}
        \vspace{-1em}
        \caption{MRPC.} 
        \label{fig:detail-mrpc}
    \end{subfigure}
    \hfill
    \caption{
        Detailed results for the English sentence classification tasks. 
        The error bars are the minimum and maximum results of the 10 random initializations.
    }
\end{figure*}

\begin{table*}[!htb]
\centering \small
\begin{tabular}{rllllll}
\toprule
Model/ target dimension & STS12 & STS13 & STS14 & STS15 & STS16 & Avg. \\
\midrule
SimCSE~\citep{gao-etal-2021-simcse} & 72.9 & \textbf{84.0} & 75.6 & 84.8 & 81.9 & 79.8 \\
CPT$_{\mathrm{XL}}$~\citep{neelakantan2022text} & 64.1 & 67.5 & 68.4 & 76.7 & 78.7 & 71.1 \\
\midrule
SBert MPNet v2 & \textbf{73.84} & 83.29 & \textbf{81.10} & \textbf{84.96} & \textbf{83.56} & \textbf{81.35} \\
sigmoid & 73.09 & 81.59 & 79.96 & 83.91 & 82.15 & 80.14 \\
256 & 70.93 & 78.92 & 77.97 & 81.93 & 80.70 & 78.09 \\
384 & 71.68 & 80.57 & 78.99 & 82.65 & 81.71 & 79.12 \\
512 & 72.28 & 81.29 & 79.41 & 83.26 & 82.13 & 79.68 \\
768 & 72.64 & 81.91 & 79.95 & 83.86 & 82.69 & 80.21 \\
1024 & 72.99 & 82.07 & 80.34 & 84.09 & 82.85 & 80.47 \\
1536 & 73.36 & 82.45 & 80.53 & 84.48 & 83.21 & 80.80 \\
2048 & 73.50 & 82.59 & 80.76 & 84.49 & 83.28 & 80.93 \\
\midrule 
SBert MiniLM v2 & 72.31 & 78.19 & 79.21 & 83.23 & 81.77 & 78.94 \\
sigmoid & 71.73 & 76.93 & 77.61 & 82.01 & 80.48 & 77.75 \\
256 & 69.71 & 74.04 & 75.95 & 79.93 & 78.84 & 75.70 \\
384 & 70.48 & 75.23 & 77.15 & 80.96 & 79.76 & 76.72 \\
512 & 71.15 & 75.77 & 77.57 & 81.63 & 80.25 & 77.27 \\
768 & 71.33 & 76.82 & 78.36 & 82.12 & 80.83 & 77.89 \\
1024 & 71.60 & 76.76 & 78.35 & 82.32 & 80.96 & 78.00 \\
1536 & 71.91 & 77.38 & 78.78 & 82.77 & 81.22 & 78.41 \\
2048 & 72.01 & 77.61 & 78.78 & 82.85 & 81.27 & 78.50 \\
\midrule 
SBert DistilUSE v2 & 69.32 & 71.68 & 74.43 & 82.60 & 80.14 & 75.63 \\
sigmoid & 67.76 & 69.21 & 73.39 & 81.09 & 78.55 & 74.00 \\
256 & 66.16 & 65.95 & 70.48 & 78.41 & 76.60 & 71.52 \\
384 & 66.87 & 67.63 & 71.81 & 79.50 & 77.64 & 72.69 \\
512 & 67.37 & 68.50 & 72.48 & 80.00 & 78.13 & 73.30 \\
768 & 68.00 & 69.63 & 73.05 & 80.77 & 78.67 & 74.03 \\
1024 & 68.30 & 69.98 & 73.22 & 81.44 & 79.23 & 74.43 \\
1536 & 68.59 & 70.62 & 73.73 & 81.74 & 79.26 & 74.79 \\
2048 & 68.91 & 70.83 & 73.99 & 82.04 & 79.57 & 75.07 \\
\midrule 
LaBSE & 67.85 & 69.10 & 71.38 & 79.89 & 74.93 & 72.63 \\
sigmoid & 67.91 & 65.93 & 70.13 & 78.25 & 73.91 & 71.23 \\
256 & 63.61 & 62.29 & 66.56 & 74.14 & 71.10 & 67.54 \\
384 & 64.12 & 64.92 & 67.92 & 75.76 & 72.59 & 69.06 \\
512 & 65.39 & 65.52 & 68.74 & 76.75 & 72.63 & 69.81 \\
768 & 66.31 & 67.26 & 69.72 & 77.78 & 73.76 & 70.97 \\
1024 & 66.63 & 67.18 & 69.91 & 77.98 & 74.01 & 71.14 \\
1536 & 67.39 & 67.64 & 70.39 & 78.78 & 74.18 & 71.68 \\
2048 & 67.37 & 67.94 & 70.67 & 78.97 & 74.15 & 71.82 \\
\midrule 
m-USE & 69.98 & 73.10 & 74.71 & 81.84 & 79.79 & 75.88 \\
sigmoid & 68.44 & 70.74 & 72.31 & 79.29 & 78.29 & 73.82 \\
256 & 66.02 & 67.31 & 70.33 & 77.08 & 76.64 & 71.48 \\
384 & 67.30 & 68.98 & 71.90 & 78.85 & 77.41 & 72.89 \\
512 & 68.14 & 69.89 & 72.43 & 79.42 & 77.94 & 73.56 \\
768 & 68.60 & 70.63 & 73.06 & 80.17 & 78.77 & 74.24 \\
1024 & 68.59 & 71.32 & 73.50 & 80.60 & 78.84 & 74.57 \\
1536 & 69.08 & 72.03 & 73.92 & 80.96 & 79.21 & 75.04 \\
2048 & 69.43 & 72.30 & 74.05 & 81.08 & 79.48 & 75.27 \\
\midrule 
LASER en & 62.52 & 59.95 & 67.72 & 76.45 & 72.89 & 67.91 \\
sigmoid & 51.12 & 50.60 & 55.66 & 61.16 & 55.88 & 54.88 \\
256 & 56.63 & 51.81 & 60.43 & 68.04 & 66.67 & 60.72 \\
384 & 58.86 & 54.50 & 62.32 & 70.69 & 68.84 & 63.04 \\
512 & 59.32 & 56.63 & 64.22 & 71.52 & 69.62 & 64.26 \\
768 & 60.28 & 56.91 & 65.01 & 73.25 & 70.14 & 65.12 \\
1024 & 61.10 & 57.90 & 65.68 & 73.77 & 71.09 & 65.91 \\
1536 & 61.47 & 58.40 & 66.28 & 74.82 & 71.85 & 66.56 \\
2048 & 61.57 & 58.84 & 66.69 & 75.04 & 71.91 & 66.81 \\
\bottomrule
\end{tabular}
\caption{Avg. Spearman correlation on English STS tasks from Senteval using different HRP layer sizes. The reported scores are the average of the 10 runs with random HRP initializations. SimCSE reports the scores of a recent state-of-the-art model.
} 
\label{tab:senteval-sts}
\end{table*}

\begin{figure*}[!ht]
    \begin{subfigure}[b]{0.49\textwidth}
        \centering
        \includegraphics[width=.95\textwidth]{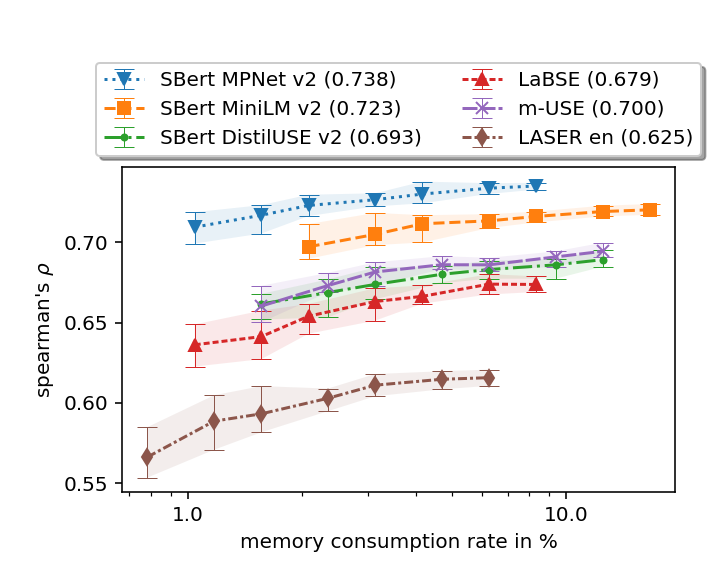}
        \vspace{-1em}
        \caption{STS12.} 
        \label{fig:detail-sts12}
    \end{subfigure}
    \hfill
    \begin{subfigure}[b]{0.49\textwidth}
        \centering
        \includegraphics[width=.95\textwidth]{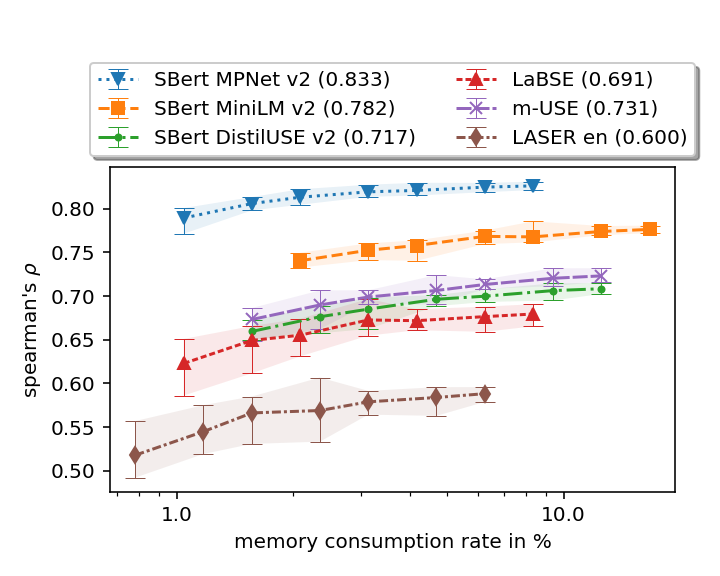}
        \vspace{-1em}
        \caption{} 
        \label{fig:detail-sts13}
    \end{subfigure}
    \hfill 
    \begin{subfigure}[b]{0.49\textwidth}
        \centering
        \includegraphics[width=.95\textwidth]{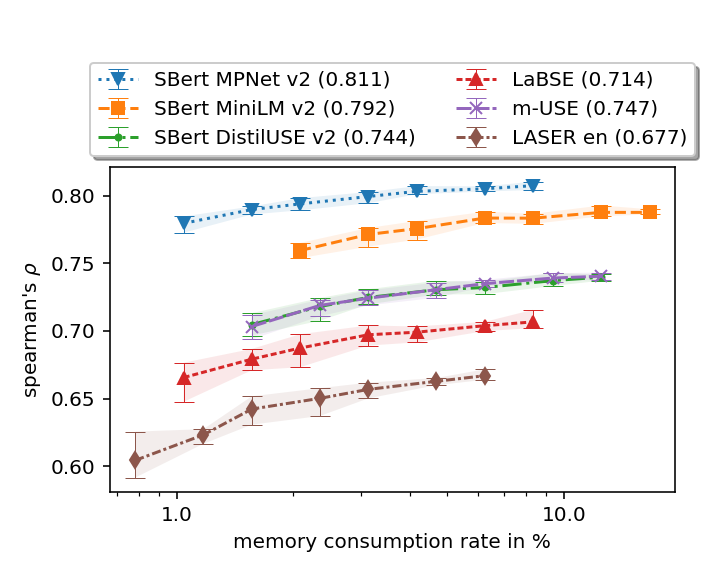}
        \vspace{-1em}
        \caption{STS12.} 
        \label{fig:detail-sts14}
    \end{subfigure}
    \hfill
    \begin{subfigure}[b]{0.49\textwidth}
        \centering
        \includegraphics[width=.95\textwidth]{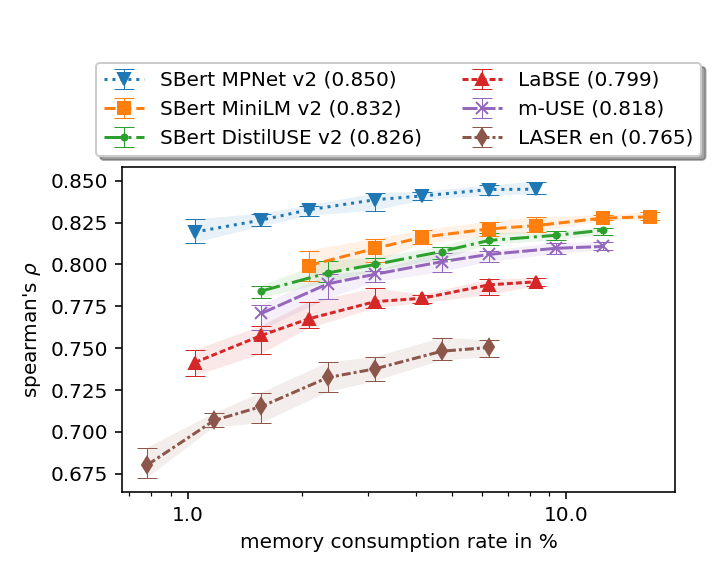}
        \vspace{-1em}
        \caption{STS15} 
        \label{fig:detail-sts15}
    \end{subfigure}
    \hfill 
    \begin{subfigure}[b]{0.49\textwidth}
        \centering
        \includegraphics[width=.95\textwidth]{img/res-sts-memorate-error-STS12}
        \vspace{-1em}
        \caption{STS16.} 
        \label{fig:detail-sts16}
    \end{subfigure}
    \hfill
    \caption{
        Detailed results for the English sentence similarity tasks. 
        The error bars are the minimum and maximum results of the 10 random initializations.
    }
\end{figure*}


\begin{table*}[!htb]
\centering \small
\begin{tabular}{rllllll}
\toprule
Model/ target dimension & VMWE & ABSD-2 & MIO-P & ARCHI & Avg. \\
\midrule
ELMo LSTM+MLP~\citep{ehren-etal-2020-supervised} & 90.1 & - & - & - & - \\
SWN2-RNN~\citep{Naderalvojoud2018GermEval2017SB} & - & 74.9 & - & - & - \\
Doc2Vec~\citep{Schabus2017} & - & - & 70.63$^{F_1}$ & - & - \\
SVM~\citep{malmasi-zampieri-2017-german} & - & - & - & 66.2$^{F_1}$ & - \\
\midrule 
SBert MPNet v2 & 83.07 & 66.54 & 89.70 & 42.23 & 70.38 \\
sigmoid & 81.96 & 65.44 & 89.59 & 36.81 & 68.45 \\
256 & 80.58 & 61.71 & 86.33 & 35.82 & 66.11 \\
384 & 81.15 & 63.10 & 87.10 & 36.45 & 66.95 \\
512 & 81.45 & 64.44 & 88.35 & 36.26 & 67.63 \\
768 & 82.18 & 65.89 & 89.19 & 37.00 & 68.57 \\
1024 & 83.05 & 66.63 & 89.93 & 36.77 & 69.09 \\
1536 & 83.48 & 68.28 & 90.58 & 36.93 & 69.82 \\
2048 & 83.88 & 69.53 & 91.15 & 37.39 & \textbf{70.49} \\
\midrule 
SBert MiniLM v2 & 83.00 & 63.29 & 87.79 & 37.66 & 67.93 \\
sigmoid & 82.31 & 62.39 & 86.83 & 33.71 & 66.31 \\
256 & 81.12 & 59.93 & 84.69 & 32.81 & 64.64 \\
384 & 82.09 & 61.86 & 86.00 & 33.94 & 65.97 \\
512 & 81.83 & 62.68 & 87.13 & 34.16 & 66.45 \\
768 & 82.68 & 64.11 & 88.09 & 34.60 & 67.37 \\
1024 & 82.98 & 65.94 & 89.07 & 34.75 & \textbf{68.18} \\
1536 & 83.72 & 67.71 & 89.87 & 35.03 & \textbf{69.08} \\
2048 & 84.64 & 69.24 & 90.31 & 35.30 & \textbf{69.87} \\
\midrule 
SBert DistilUSE v2 & 82.24 & 64.19 & 90.21 & 38.60 & 68.81 \\
sigmoid & 82.31 & 60.90 & 90.21 & 36.96 & 67.59 \\
256 & 79.48 & 58.22 & 88.34 & 34.16 & 65.05 \\
384 & 80.68 & 59.86 & 89.56 & 35.18 & 66.32 \\
512 & 82.13 & 61.55 & 90.06 & 34.93 & 67.17 \\
768 & 82.09 & 62.97 & 90.72 & 35.65 & 67.86 \\
1024 & 82.63 & 64.02 & 91.12 & 36.00 & 68.44 \\
1536 & 83.24 & 66.64 & 91.58 & 36.33 & \textbf{69.45} \\
2048 & 84.08 & 67.33 & 91.87 & 36.30 & \textbf{69.90} \\
\midrule 
LaBSE & 83.41 & 65.40 & 90.21 & 38.54 & 69.39 \\
sigmoid & 82.45 & 67.05 & 90.98 & 36.37 & 69.21 \\
256 & 80.70 & 60.69 & 88.17 & 33.87 & 65.86 \\
384 & 81.40 & 61.39 & 89.11 & 34.63 & 66.63 \\
512 & 82.01 & 63.01 & 89.94 & 35.60 & 67.64 \\
768 & 83.03 & 64.62 & 90.95 & 35.18 & 68.44 \\
1024 & 83.12 & 66.22 & 91.26 & 35.47 & 69.02 \\
1536 & 83.95 & 68.13 & 91.74 & 35.53 & \textbf{69.84} \\
2048 & 84.50 & 68.79 & 92.17 & 35.87 & \textbf{70.33} \\
\midrule 
m-USE & 84.38 & 62.50 & 89.76 & 44.44 & 70.27 \\
sigmoid & 81.06 & 61.10 & 89.10 & 39.72 & 67.74 \\
256 & 79.39 & 57.59 & 86.90 & 36.95 & 65.21 \\
384 & 80.11 & 59.11 & 88.04 & 38.34 & 66.40 \\
512 & 81.03 & 60.32 & 88.85 & 38.80 & 67.25 \\
768 & 81.11 & 62.21 & 89.92 & 39.52 & 68.19 \\
1024 & 81.78 & 63.31 & 90.27 & 39.68 & 68.76 \\
1536 & 82.79 & 65.51 & 91.02 & 39.28 & 69.65 \\
2048 & 83.17 & 66.43 & 91.16 & 38.76 & 69.88 \\
\midrule 
LASER de & 85.28 & 65.56 & 90.25 & 46.22 & 71.83 \\
sigmoid & 82.31 & 54.01 & 87.51 & 38.18 & 65.50 \\
256 & 76.50 & 53.48 & 84.97 & 36.18 & 62.78 \\
384 & 77.35 & 55.48 & 86.94 & 37.54 & 64.33 \\
512 & 78.37 & 56.86 & 87.85 & 38.29 & 65.34 \\
768 & 78.85 & 59.38 & 89.18 & 39.12 & 66.63 \\
1024 & 79.67 & 61.67 & 90.06 & 39.40 & 67.70 \\
1536 & 80.10 & 63.74 & 90.81 & 39.98 & 68.66 \\
2048 & 81.85 & 64.65 & 91.26 & 40.35 & 69.53 \\
\bottomrule
\end{tabular}
\caption{Average accuracy for sentence embedding models using HRP-layers on German SEEG classification tasks. $F_1$ indicates cases where only the F1 score was reported for previous models. All scores are averaged across 10 runs with random HRP initializations and different random seeds. \\
} 
\label{tab:seeg-downstream}
\end{table*}

\begin{figure*}[!ht]
    \begin{subfigure}[b]{0.49\textwidth}
        \centering
        \includegraphics[width=.95\textwidth]{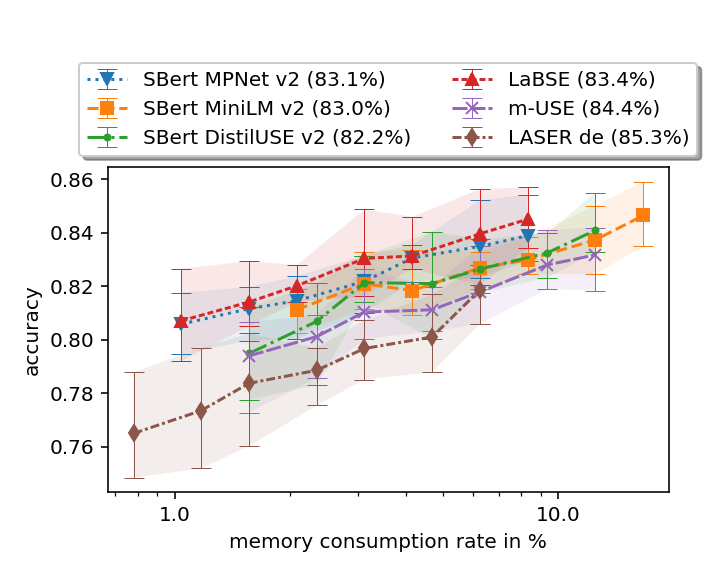}
        \vspace{-1em}
        \caption{VMWE.} 
        \label{fig:detail-vmwe}
    \end{subfigure}
    \hfill  
    \begin{subfigure}[b]{0.49\textwidth}
        \centering
        \includegraphics[width=.95\textwidth]{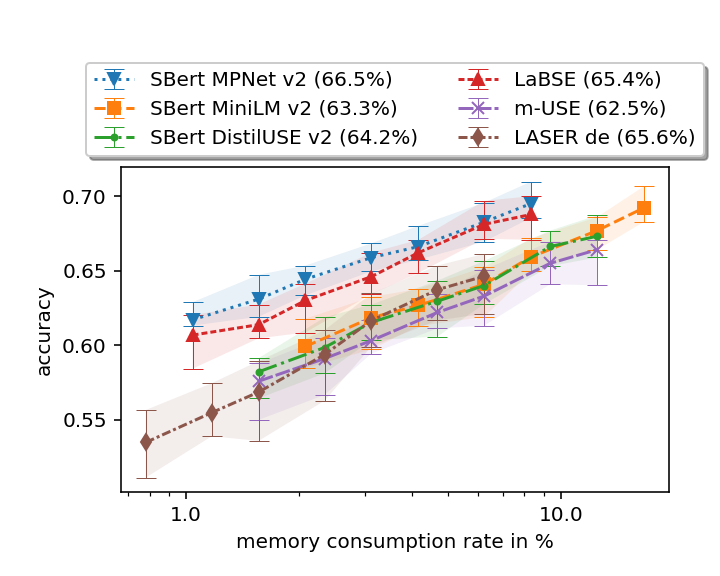}
        \vspace{-1em}
        \caption{ABSD-2.} 
        \label{fig:detail-absd2}
    \end{subfigure}
    \hfill
    \begin{subfigure}[b]{0.49\textwidth}
        \centering
        \includegraphics[width=.95\textwidth]{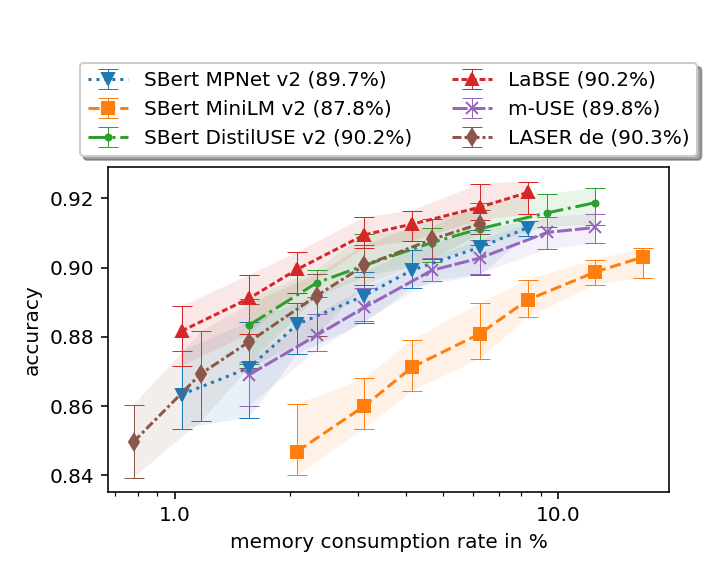}
        \vspace{-1em}
        \caption{MIO-P.} 
        \label{fig:detail-miop}
    \end{subfigure}
    \hfill
    \begin{subfigure}[b]{0.49\textwidth}
        \centering
        \includegraphics[width=.95\textwidth]{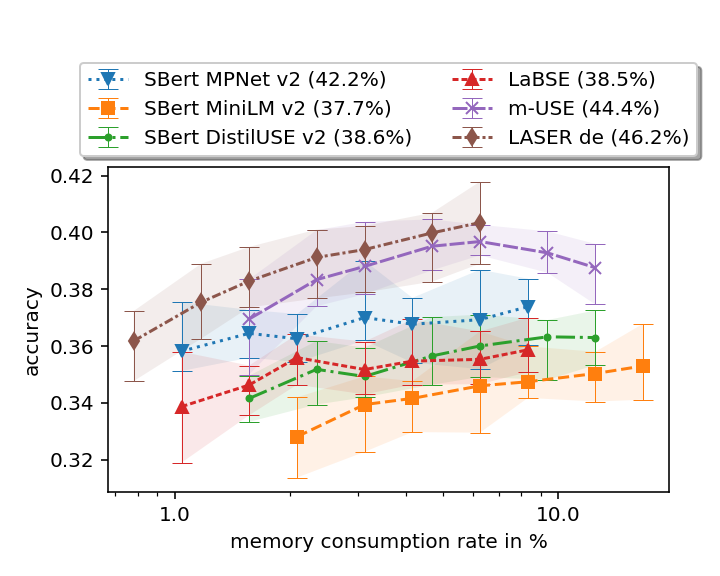}
        \vspace{-1em}
        \caption{ARCHI.} 
        \label{fig:detail-archi}
    \end{subfigure}
    \hfill 
    \caption{
        Detailed results for the German sentence classification tasks. The results of the original 
        The error bars are the minimum and maximum results of the 10 random initializations.
    }
\end{figure*}

\end{document}